\def\year{2021}\relax
\documentclass[letterpaper]{article} 
\usepackage{aaai21}  
\usepackage{times}  
\usepackage{helvet} 
\usepackage{courier}  
\usepackage[hyphens]{url}  
\usepackage{graphicx} 
\usepackage{natbib}  
\usepackage{amsmath}
\usepackage{caption} 
\usepackage{todonotes}
\usepackage{soul}  

\title{Data-based Discovery of Governing Equations
 }
\author{
	Waad Subber\textsuperscript{\rm 1}\thanks{Corresponding Author: Waad Subber, email: Waad.Subber@ge.com},
  Piyush Pandita\textsuperscript{\rm 1},  
  Sayan Ghosh\textsuperscript{\rm 1},
  Genghis Khan\textsuperscript{\rm 1},      
  Liping Wang\textsuperscript{\rm 1},
  Roger Ghanem\textsuperscript{\rm 2}
  \\
}
\affiliations{
    \textsuperscript{\rm 1}GE Research, Niskayuna, NY 12309\\
    \textsuperscript{\rm 2}University of Southern California, Los Angeles, CA 90007

}

\begin{document}

\maketitle

\begin{abstract}
Most common mechanistic models are traditionally presented in mathematical forms to explain a given physical phenomenon. 	Machine learning algorithms, on the other hand, provide a mechanism to map the input data to output without explicitly  describing the underlying physical process that generated the data. We propose a Data-based Physics Discovery (DPD) framework for  automatic discovery of governing equations from observed data. Without a prior definition of the model structure, first a free-form of the equation is discovered, and then calibrated and validated against the available data. In addition to the observed data, the DPD framework can utilize available prior physical models, and  domain expert feedback. When prior models are available, the DPD framework can discover an additive or multiplicative correction term represented symbolically. The correction term can be a function of the existing input variable to the prior model, or a newly introduced variable. In case a prior model is not available, the DPD framework discovers a new data-based standalone model governing the observations. We demonstrate the performance of the proposed framework on a real-world application in the aerospace industry.
\end{abstract}

\section{Introduction}

Modern machine learning (ML) methods are aimed at providing a statistical mechanism to predict the outcome of a system under new conditions. 
This statistical mechanism is constructed based on exploring the correlation between inputs and outputs that is embedded in data~\cite{jain2003water}. 
However, in many engineering applications, the inputs, outputs, and ML model structure are not selected such that learning elucidates insight into the underlying physical process that generated the data beyond a black-box function approximation. 
Thus, knowledge is discovered in a data-driven manner without fully explaining the physics of the problem. 
The mechanistic modeling approach, on the other hand, proceeds from a starting point of scientific laws and axioms, producing by way of logical deduction formal models of the physics underlying a phenomenon and measurements of it. Typically, mechanistic modeling approach describes causal mechanisms by simplified mathematical formulations, while the ML approach seeks to establish a statistical relationship between inputs and outputs. These two approaches  should not be seen as direct competitors~\cite{baker2018mechanistic}. 
The advantages of one approach should be used to complement its counterpart, which suggests that modern research efforts in the scientific machine learning field should be directed towards enabling a symbiotic relationship between both approaches~\cite{baker2019workshop,jain2003water,baker2018mechanistic}. A synergy framework between machine learning and mechanistic approach can be built based on integrating multiple sources of information (such as field and lab data), prior domain knowledge, physical constraints and expert feedback in one unified framework. The main feature of such an approach is manifested in the symbolic representation of the predictive model. The symbolic description of the prediction mechanism by mathematical expressions can provide explainability for its predictions, facilitate integrating expert feedback, and fuse the state-of-art domain knowledge in an explicit manner.

To this end, we  propose a Data-based Physics Discovery (DPD) framework for  automatic discovery of governing equations from observed data. Symbolic regression and Bayesian calibration are utilized in discovering the physical laws governing the data. The approach is based on integrating  multiple sources of data, domain knowledge, physical constraints and expert feedback in one unified framework for model discovery. In our previous work \cite{atkinson2019data}, we introduced a method to infer a symbolic representations of differential operators  from data, in a free-form manner as opposed to methods such as SINDy \cite{brunton2016discovering} which require a user to postulate a library of terms from which only linear combinations might be considered. In this work, we propose a framework to  discover governing equation from data utilizing prior physical  domain knowledge and constraints. In case a prior physical model is available, the DPD framework discovers a multiplicative or additive correction term; otherwise a standalone model based on data only is proposed. The discovered correction term or standalone model can be a function of existing input variable or a newly introduced variable. The symbolic representation of the discovered model helps in explaining the effect of a new variable on the current physical process. The discovered  model is calibrated using Bayesian Hybrid Modeling (GEBHM)~ approach \cite{ghosh2020advances, zhang2020remarks}. The GEBHM is a probabilistic ML method that enables calibration, validation, multi-fidelity modeling and uncertainty quantification.  In addition, the framework equipped with an optimal experiment design tool to propose  a new experiment for enhancing the model accuracy. Here, the Intelligent Design and Analysis of Computer Experiments (IDACE)~\cite{kristensen2019industrial} methodology is utilized. The IDACE is a technique that adaptively generates new experiment setting for improving the accuracy of the model based on the estimated  uncertainty and output desirability.  
 
We provide a technical description of the approach and demonstrate its use on a real-world problem with relevance to the aerospace industry. The  application is focused on discovering a predictive model describing corrosion, given the availability of a prior physical model. We demonstrate the ability of the proposed framework to fuse prior knowledge with a discovered correction term, improving the model's fidelity to measurements.

\section{Data-based Physics Discovery (DPD) framework }
\label{DPD}
\begin{figure}[t]
\centering
\includegraphics[width=0.9\columnwidth]{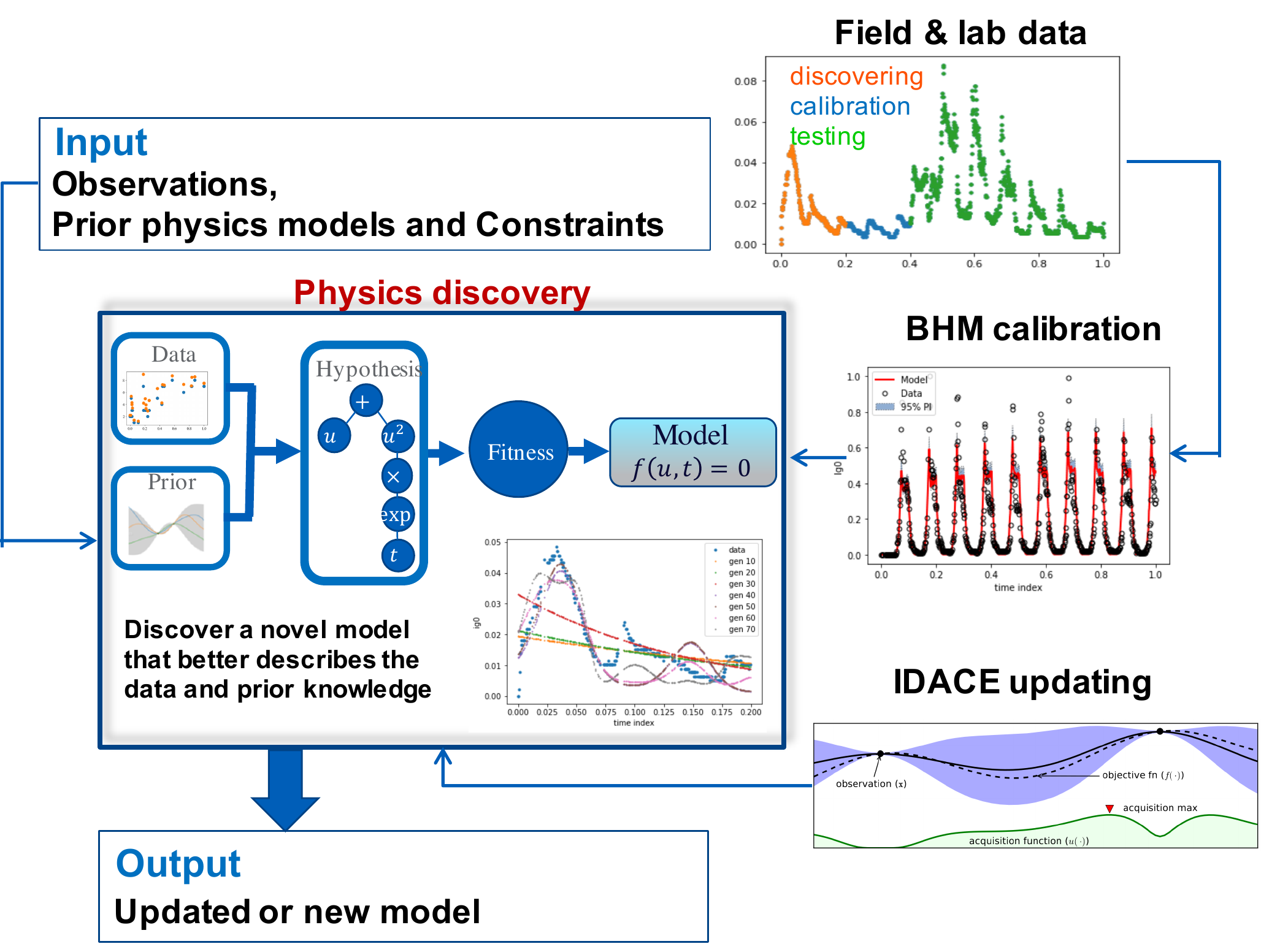} 
\caption{The workflow of the Data-based Physics Discovery (DPD) framework. The inputs to the framework are observation, and optional prior model and constraints. First, depending on the setting, a standalone model or a correction term is discovered in the physical discovery part. Second, the discovered model is calibrated. Third, V\&V and UQ are performed to the  discovered model and, if needed, a new set of experiments are proposed to improve the accuracy. The output is a calibrated and validated model.}
\label{workflow}
\end{figure}

The proposed Data-based Physics Discovery (DPD)   approach for integrating domain knowledge, physical constraints and observational data in one framework is developed to predict the response (with confidence bounds) of complex engineering systems under novel conditions. 
This framework enables a  relationship between data-based machine learning and causality-based mechanistic modeling approaches. The DPD framework serves as a human’s research assistant with two goals in mind: enhance the credibility of prediction and enable learning new physics.
This constitutes a new model-building paradigm whereby machine learning algorithms and scientists work together to formulate a symbolic mathematical form explaining a novel physical phenomenon. 
Similar to the physical experiments performed by scientists to discover the interaction relationship between the input and output variables, we propose a set of variables to our symbolic regression framework and ask for a relationship govern them. 
An overview  for the workflow of the proposed approach is shown in Fig.~(\ref{workflow}).

Given set of observations ${\mathcal D}={\{\bf x}_i, {\bf y}_i\}_{i=1}^n$,  (and  if available, a prior physical model  ${\bf u} = f({\bf x},\theta_p)$ and a constraint  $h({\bf x}) \geq 0$),  we seek  a new or updated model that better describes the observed data.  The process as shown in Fig.~(\ref{workflow}) can be summarized as follows:
{\bf(i)} Let a prior model $f({\bf x},\theta_p)$, where $\theta_p$ are the calibration parameters, be available and an additive correction term is required, then the problem is posed as: given ${\mathcal D}$ and $f({\bf x},\theta_p)$, find $\delta({\bf x},\theta_d)$ such that  $\|g({\bf x},\theta)-{\bf y}\|_2 $ is minimized, where $g({\bf x,\theta})= f({\bf x},\theta_p) +\rho \delta({\bf x},\theta_d)$.  Here $\theta=\{\theta_p,\theta_d,\rho\}$,  $\rho$ is a weighting parameter, and $\theta_d$ represent the ephemeral random constants generated during the evolution of the symbolic regression~\cite{DEAP_JMLR2012}. This setting can be readily generalized to the case of multiplicative correction $g({\bf x},\theta)= f({\bf x},\theta_p) \times \rho \delta({\bf x},\theta_d)$, and to the case when a prior model  is not available  $g({\bf x},\theta)= \delta({\bf x},\theta_d)$. Note that we do not {\it a priori} define  a model structure of the correction term $\delta({\bf x},\theta_d)$. In this step, the scaling parameter $\rho$ is set to unity and the  prior model parameters $\theta_p$ are set to their maximum likelihood estimation values, and $\theta_d$  is created and set during the evolutionary computation at run time~\cite{koza1992programming}, {\bf(ii)} After discovering the model structure, the model parameters $\theta=\{\theta_p,\theta_d\,\rho\}$ in the newly discovered model $g({\bf x},\theta)$ are calibrated next using Bayesian approach, {\bf(iii)} In case the calibrated model does not satisfy the V\&V  requirements (e.g., test data are within $95\%$ confidence interval of the model prediction), a new optimal experiment for enhancing the model accuracy is proposed  using an intelligent design of experiments.  Once the discovered model  satisfies the V\&V and UQ requirements it will be proposed as an explainable model to the domain experts for their feedback.

In essence, the proposed workflow embeds domain knowledge and physical laws into machine learning algorithms to provide an interpretable predictive model. Salient features of the DPD framework are: 1) enhancing the forecasting ability of the machine learning algorithms by embedding physics principles enabling extrapolation for regions were data is not available, 2) addressing the limitations in the traditional mechanistic modeling approach in incorporating multiple sources of  information, 3) minimizing the size of the training data  required for machine learning approach by incorporating domain knowledge.

\section{Real-World Application}

An aerospace industrial application is considered to demonstrate the potential practicality of the proposed framework. The problem is focused on developing a predictive model for the corrosion of an aircraft structural materials. In this application, we  demonstrate: 1) the ability of  DPD in discovering a  correction term to an existing physical model for better forecasting ability, 2) the improved performance obtained  by the DPD  compared to an existing physical model for the case of small training data set, 3)  the performance of the DPD  in discovering a standalone model based on only the training data set  without incorporating a prior physical model.

\subsection{Corrosion problem}
Extending the service lifetime of a corrosion protection coating system necessitates a fundamental understanding of the materials  under harsh operation conditions; 
the sensitivity of the corrosion process to material features and service conditions constitutes a challenge to the development of a novel corrosion-resistant materials~\cite{olajire2018recent}. 
The prediction of the corrosion under long-term operation conditions and unforeseen environmental events is a crucial process for maintenance scheduling of the aircraft structural. To this end, our proposed framework will provide a calibrated, validated and uncertainty quantified predictive model for the corrosion as a function of the environmental and operations conditions. The symbolic representation  of the discovered corrosion model enables  interpretability of the prediction model, which is required for understanding the corrosion process under novel  conditions.

\subsubsection{Problem setting:}
The experimental data is provided by the AFRL as part of the DARPA AIRA challenge problems~\cite{AIRA}. 
The measurements are represented by time-series data of the corrosion current, temperature and relative humidity are reported every 5 minutes. The objective here is to develop a forecasting model for the corrosion current, given the environmental conditions.

For this application, we pose the problem as finding a correction term to an available prior physical model as $g({T,H,\theta})= f(T,\theta_p) \times \rho \delta(H,\theta_d)$, where $T$ is the temperature,  $H$ is relative humidity, and the $\theta=\{\theta_p,\theta_d,\rho\}$ are the calibration parameters. Note that the available prior model is a function of temperature $T$ only, while the discovered term is set to be as a function of  humidity $H$ only. This setting is to show how the  DPD  framework  can introduce a new variable to an existing physical model in order to improve the forecasting accuracy. 

\begin{figure}[t]
\centering
\includegraphics[width=0.75\columnwidth]{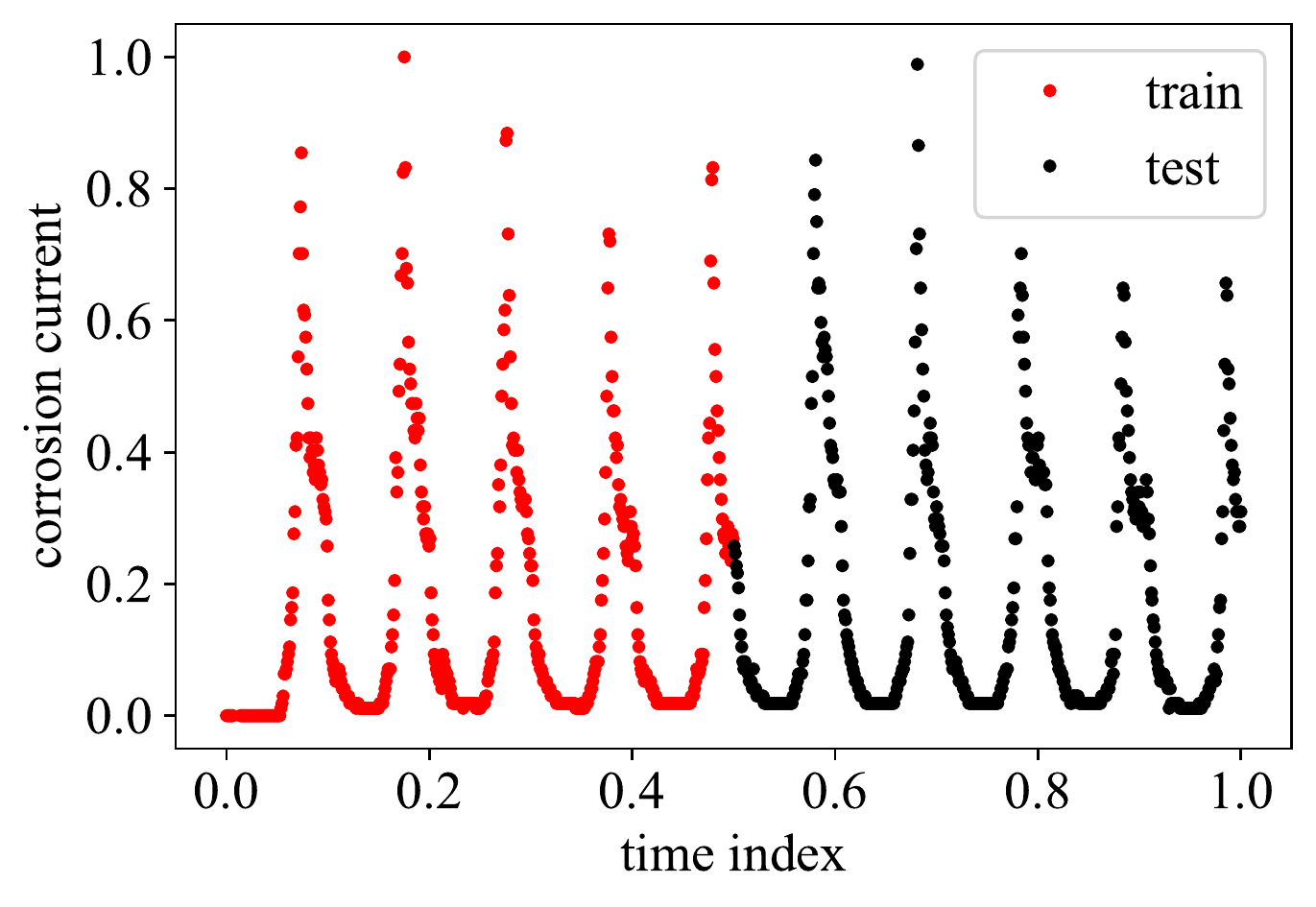} 
\caption{	The data is halved equally into training and testing sets. The decomposition is based on the time index (i.e., training data $t\le 0.5$ and testing data  $t > 0.5$).}
\label{train_test_50_50}
\end{figure}

Specifically,  the Butler-Volmer equation of the form $f(T,\theta_p)=\theta_0 \exp({\theta_1}/{T}) + \theta_2 \exp({\theta_3}/{T})$ is used as a prior model for the temperature $T$. The correction term is set as a function of  the relative humidity $H$ and is denoted by $\delta(H,\theta_d)$. The symbolic regression in the  DPD framework will discover  $\delta(H,\theta_d)$ such that  $\| g(T,H,\theta) - y \|_{2}$ is minimum, where $y$ is the measured corrosion current, $g(T,H,\theta)=f(T,\hat{\theta}_p) \times \hat{\rho} \delta(H,\theta_d)$, here $\hat{\theta}_p$ are set {\it a priori} to the maximum likelihood estimation values, and $\hat{\rho}=1$. Once the most suitable structure of the correction term is discovered, a Bayesian calibration  is performed to estimate the model parameters $\theta=\{\theta_p,\theta_d,\rho\}$. The equation discovery and the  model calibration are performed on the first $50\%$ portion of the data, whereas the prediction and testing of the model are performed on the forthcoming $50\%$ of the data as shown Fig.~(\ref{train_test_50_50}).

\subsubsection{Baseline model:}
\begin{figure}[t]
\centering
\includegraphics[width=0.75\columnwidth]{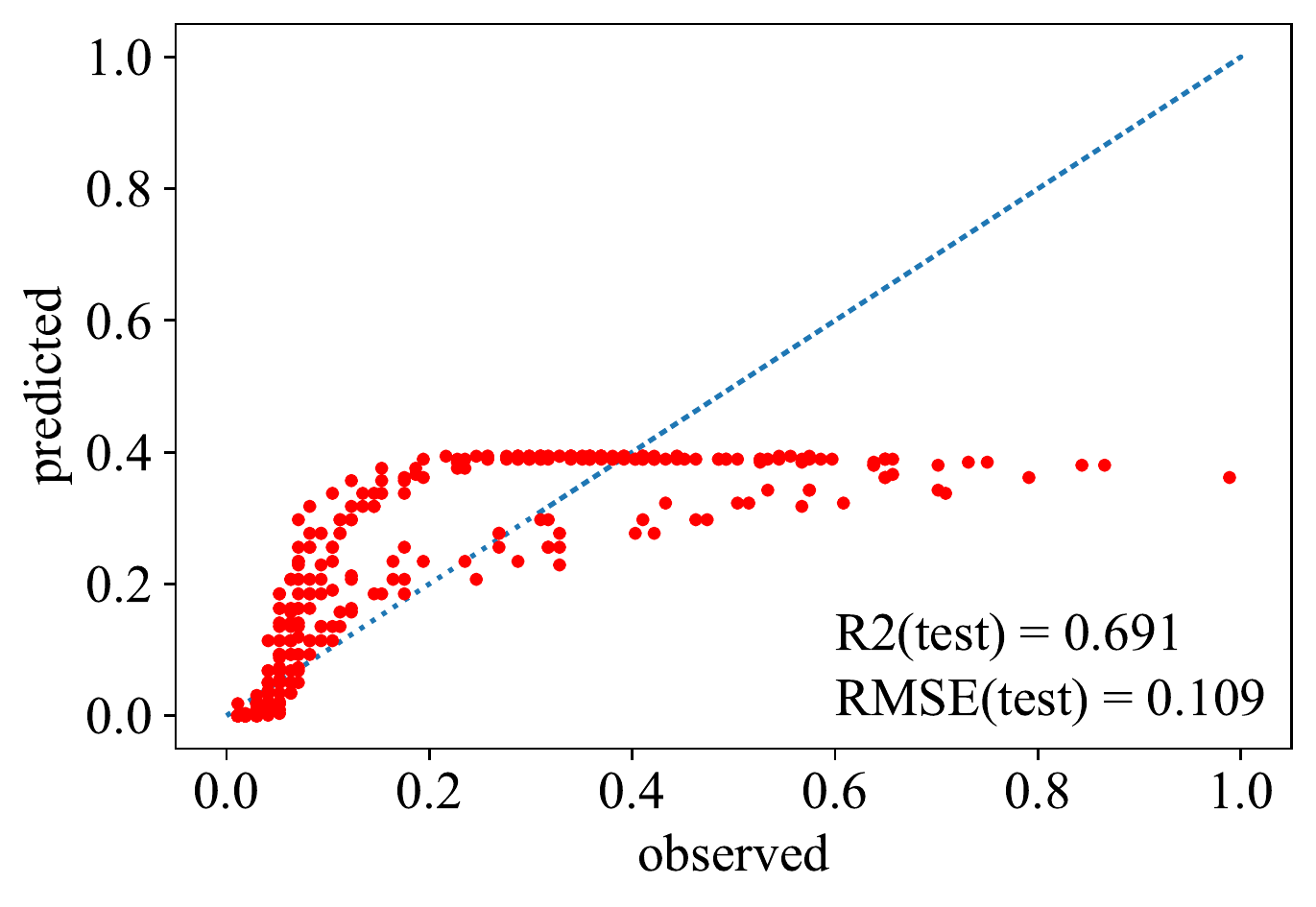} 
\caption{	Validation of the  baseline  model $f(T,\theta_p)$ on  the testing data set, $t> 0.5$. }
\label{test_deter_50_50}
\end{figure}

First, using the training data set $t\le 0.5$,  a gradient based optimization is performed to obtain  the maximum likelihood estimation (MLE) of the parameters in the Butler-Volmer equation. The model, which is a function of temperature only, is validated on the remaining test data. The validation plot for the test data is shown in Fig.~(\ref{test_deter_50_50}). The performance metrics  R-squared $R^2 = 0.691$ and root mean square error  $RMES = 0.109$ are calculated using the test data.

In order to improve  the performance of the Butler-Volmer equation without requiring extra training data, a new correction term to the existing model is introduced. The goal of the correction term is not only to enhance the model performance, but also to study the effect of newly introduced variable on the corrosion current. Commonly, the former goal is achieved by  increasing the size of the training data set, and the latter goal is carried out by a domain expert in mechanistic way. Next, we will show how DPD framework can improve the performance of an existing model by an automated discovery of a new governing equation using the  available training data set.

\subsubsection{DPD with a prior model:}
Utilizing the existing model $f(T,\hat{\theta}_p)$ with the {\it a priori} estimated parameters $\hat{\theta}_p$ and the training data set, Table~(\ref{DPD1}) lists some of the discovered multiplicative correction terms. In addition to the $R^2$ listed for each term, the Bayesian information criterion (BIC) and the gained improvement are presented. The percentage improvement is defined based on the $R^2$ value as $V[\%]=(R^2_{b}/R^2_{i} -1)\times100$, where $R^2_{b}=0.691$ is value of the baseline model and $R^2_{i}$ is the value of the new discovered model, before Bayesian calibration. The parameters $\theta_d$ of the correction term $\delta(H,\theta_d)$ are not calibrated with Bayesian approach yet.  A forecasting enhancement ranging between $10\%$-$14\%$, depending on the complexity of the obtained term, can be achieved by introducing a new variable to the existing model.

\begin{table}[t]
\caption{Discovered correction terms to the  Butler-Volmer equation for the training data set $t\le 0.5$. Performance metrics are based on the testing data set $t>0.5$.}    
\label{DPD1}
\begin{center}
    \begin{tabular}{| l | l | l | l |  }
    \hline
BIC	& $R^2$			& $\delta(H,\theta_d)$ & V[\%]\\ \hline
-2078.2	&0.787	&	$0.481H\exp(H^3)$	&14 \\ \hline
-2073.9	&0.774	&	$H(2H - 0.73)$&	12 \\ \hline
-
-2064.3	&0.763	& $2H - 0.821$ &	10 \\ \hline
    \end{tabular}
\end{center}
\end{table}

%
%

\subsubsection{Model calibration:}
\begin{figure}[t]
\centering
\includegraphics[width=0.75\columnwidth]{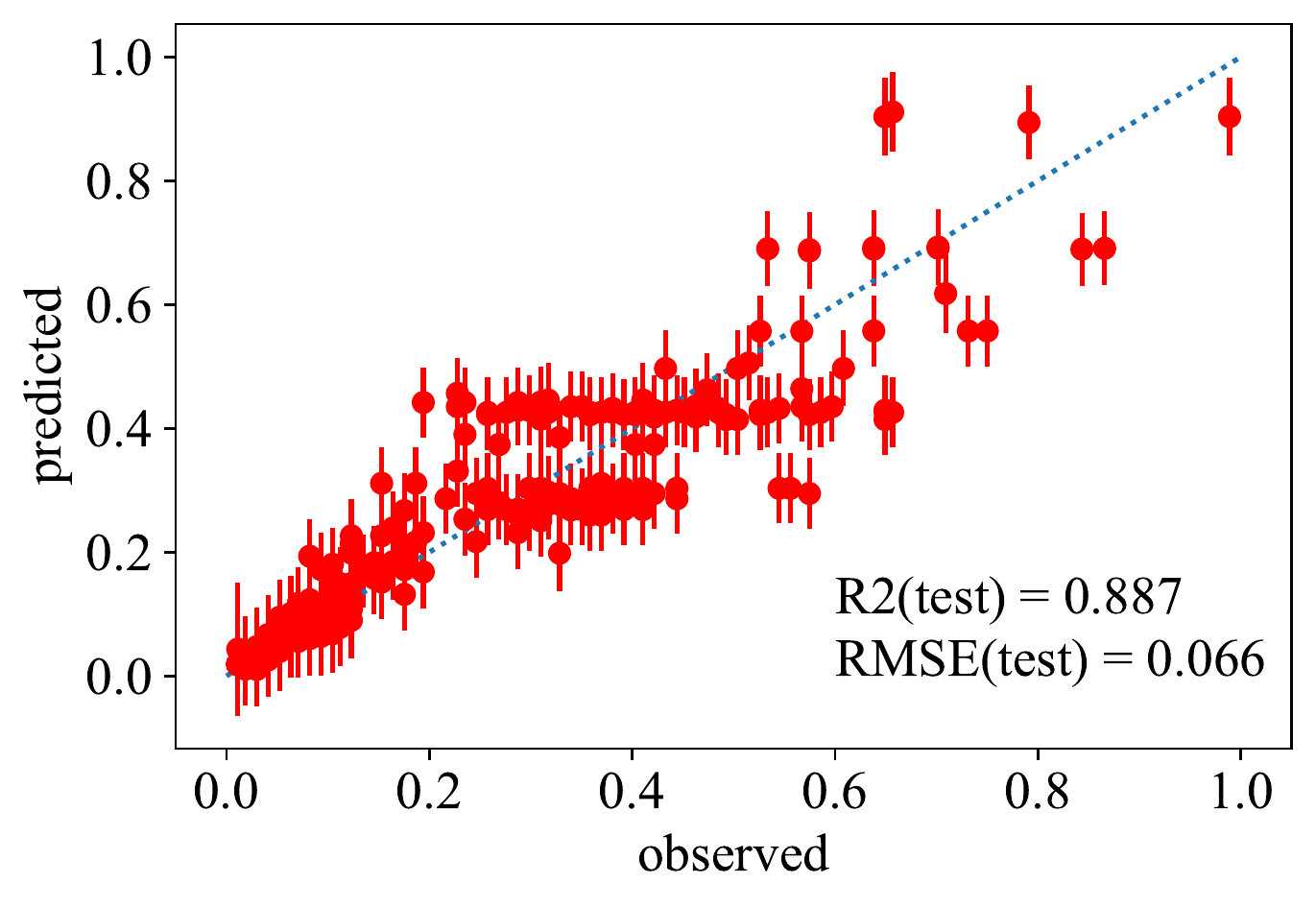} 
\caption{	Validation of the  discovered model $g(T,H,\theta)$ on the testing data set, $t> 0.5$.}
\label{test_bhm_50_50_std}
\end{figure}

\begin{figure}[t]
\centering
\includegraphics[width=0.75\columnwidth]{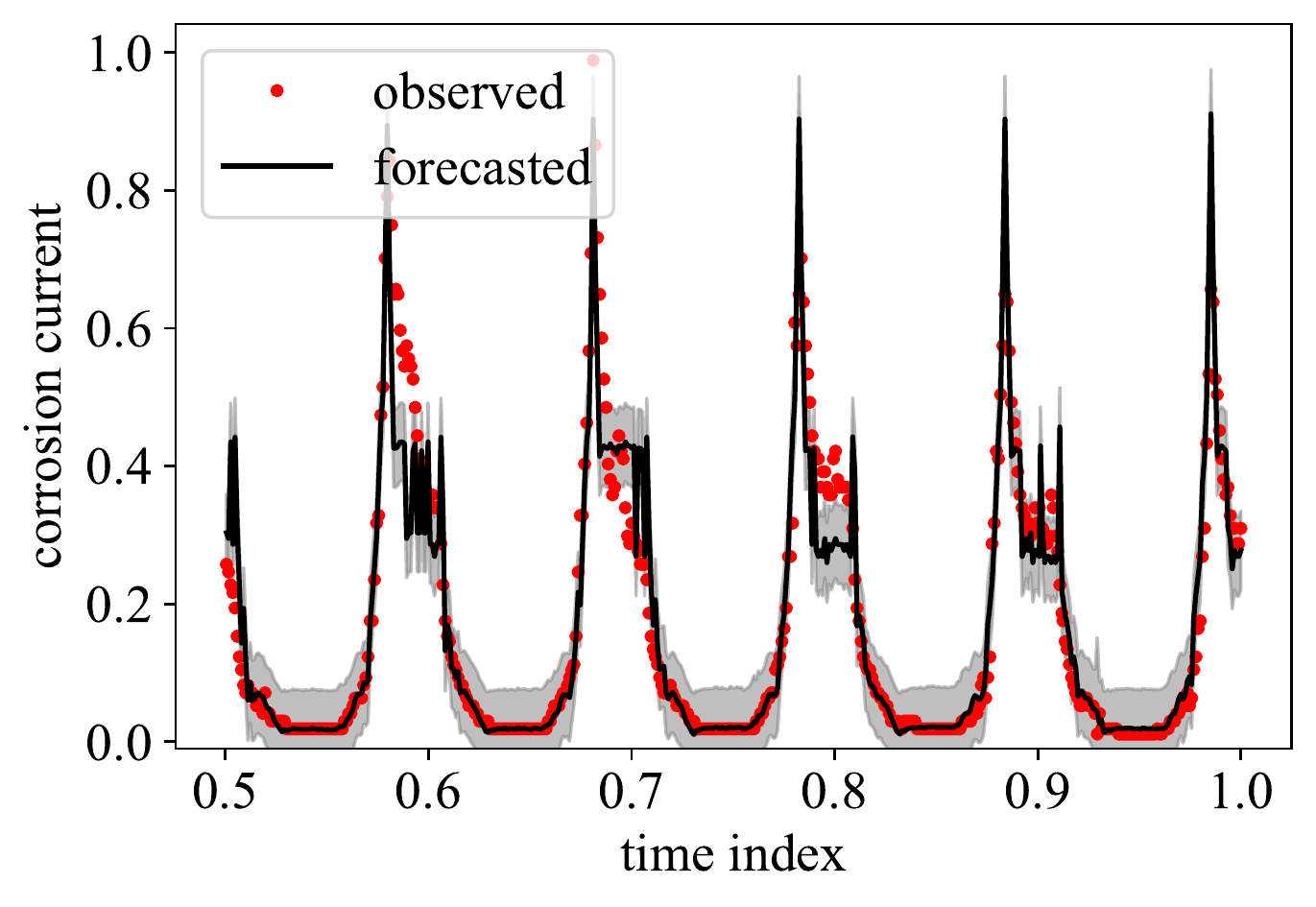} 
\caption{	The forecasting of the discovered model $g(T,H,\theta)$ for $t>0.5$. The confidence bounds are defined as ($\mu \pm \sigma$).}
\label{forecast_bhm_50_50_std}
\end{figure}

In order to capture any missing physics, the Kennedy and  O'Hagan Bayesian calibration~\cite{kennedy2001bayesian} is carried out next using the GEBHM framework. The validation results of the calibrated model is reported in  Fig.~(\ref{test_bhm_50_50_std}). An improvement of $28\%$ can be achieved by the new discovered model with parameters being calibrated using GEBHM. Fig.~(\ref{forecast_bhm_50_50_std}) shows the forecasting of the discovered model  as well as the observed data for $t>0.5$. The gray shaded area is the confidence bounds defined as $\mu \pm \sigma$.  All  data are normalized between 0 and 1.  The model prediction agrees quite well with the measured data.

\subsubsection{Reduction of the training data size:}
\begin{figure}[t]
\centering
\includegraphics[width=0.75\columnwidth]{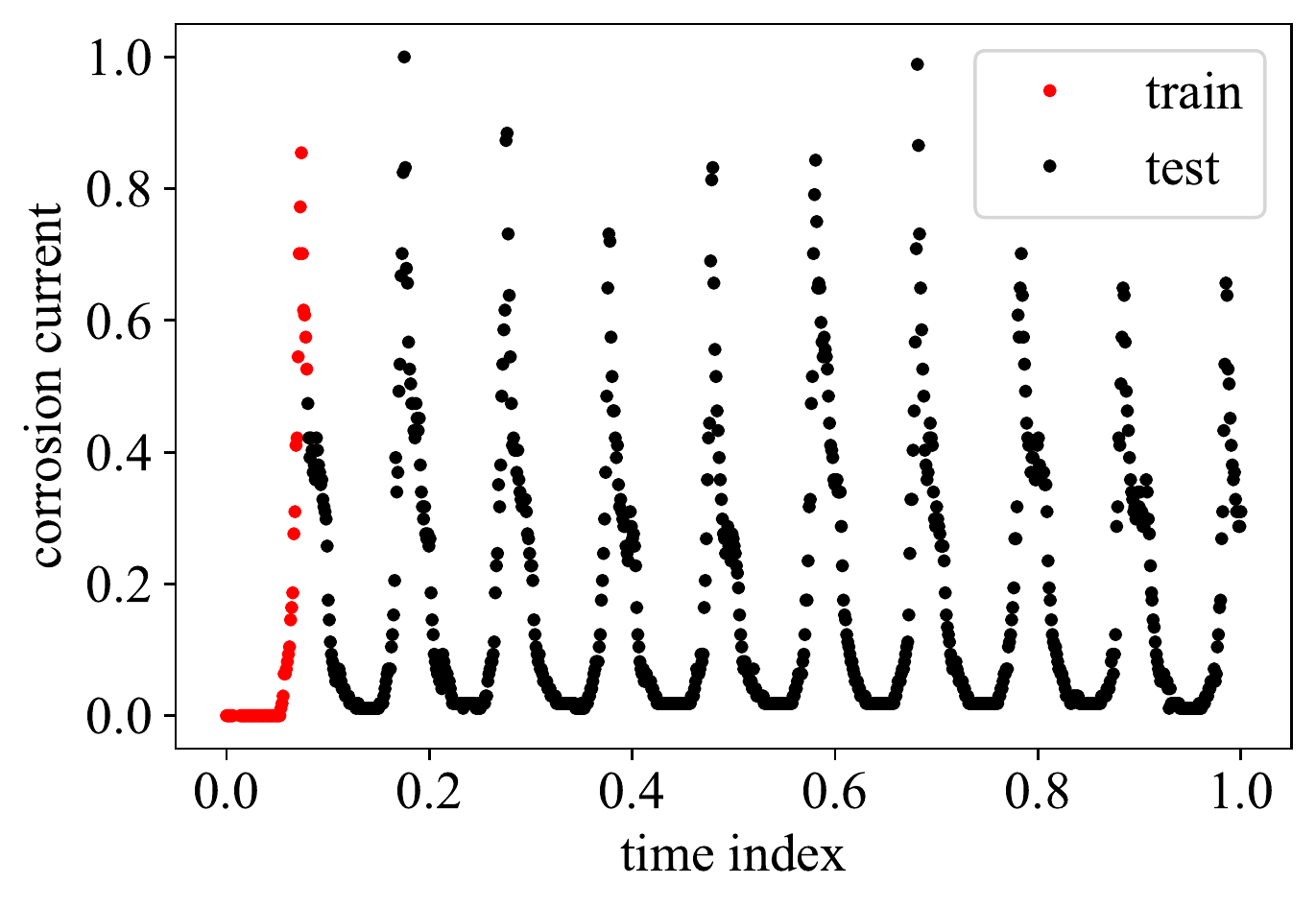} 
\caption{	The data is partitioned based on the time index into training ($t\le 0.08$) and testing  ($t > 0.08$) sets.}
\label{train_test_008}
\end{figure}

\begin{figure}[t]
\centering
\includegraphics[width=0.75\columnwidth]{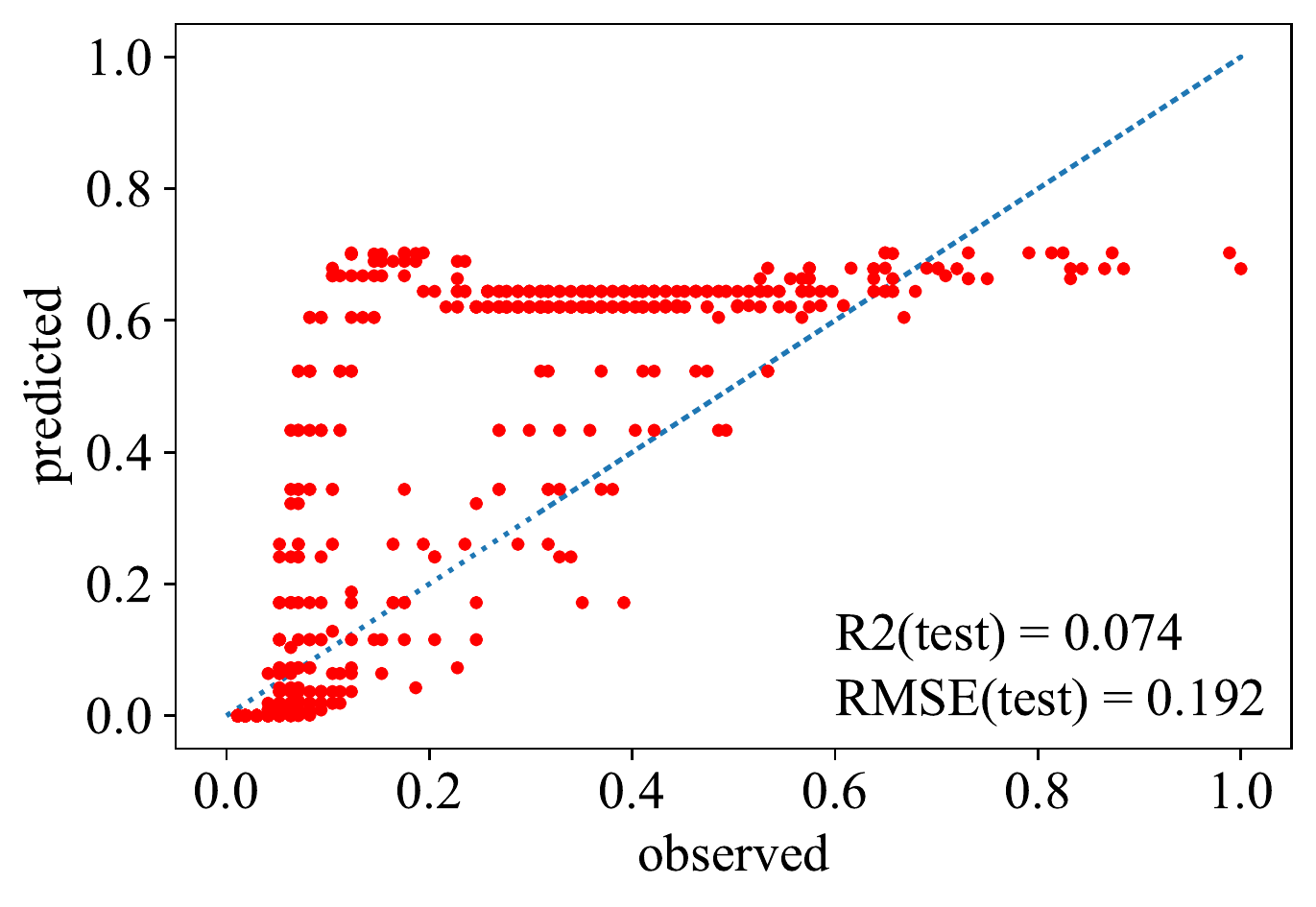} 
\caption{	Validation of the  baseline  model $f(T,{\theta}_p)$ on the testing data set, $t> 0.08$.}
\label{test_deter_008}
\end{figure}

To study the effect of the training data size on the model performance, we use only the first $0.08\%$ of the available data for training and the remaining portion are set for testing  as shown in Fig.~(\ref{train_test_008}). Fig.~(\ref{test_deter_008}) shows the validation plot of the current baseline model  (the Butler-Volmer with maximum likelihood estimation of the parameters). For the training data size  ($t \le 0.08$), the performance metrics of the baseline model are much lower than that obtained when using half of the data   ($t \le 0.5$) as shown in Fig.~(\ref{test_deter_50_50}). Precisely a reduction of $(89\%)$ in the $R^2$ value is observed. This observation highlights the effect of the training data size on the performance of the model. Next, using the available training set ($t \le 0.08$) we show an improvement can be obtained using the DPD framework. The improvement is achieved by  introducing a  correction term as a function of a new input variable.

For the given training data set ($t\le 0.08$),  Table~(\ref{DPD2}) lists the discovered correction terms  $\delta(H,\theta_d)$ with the corresponding BIC, $R^2$ and the percentage value of improvement obtained, before calibrating the model. By introducing a new variable to the existing model, the DPD discovered a correction term (e.g., $0.391\exp(H)$) that can lead to an improvement of $78\%$.

\begin{figure}[t]
\centering
\includegraphics[width=0.75\columnwidth]{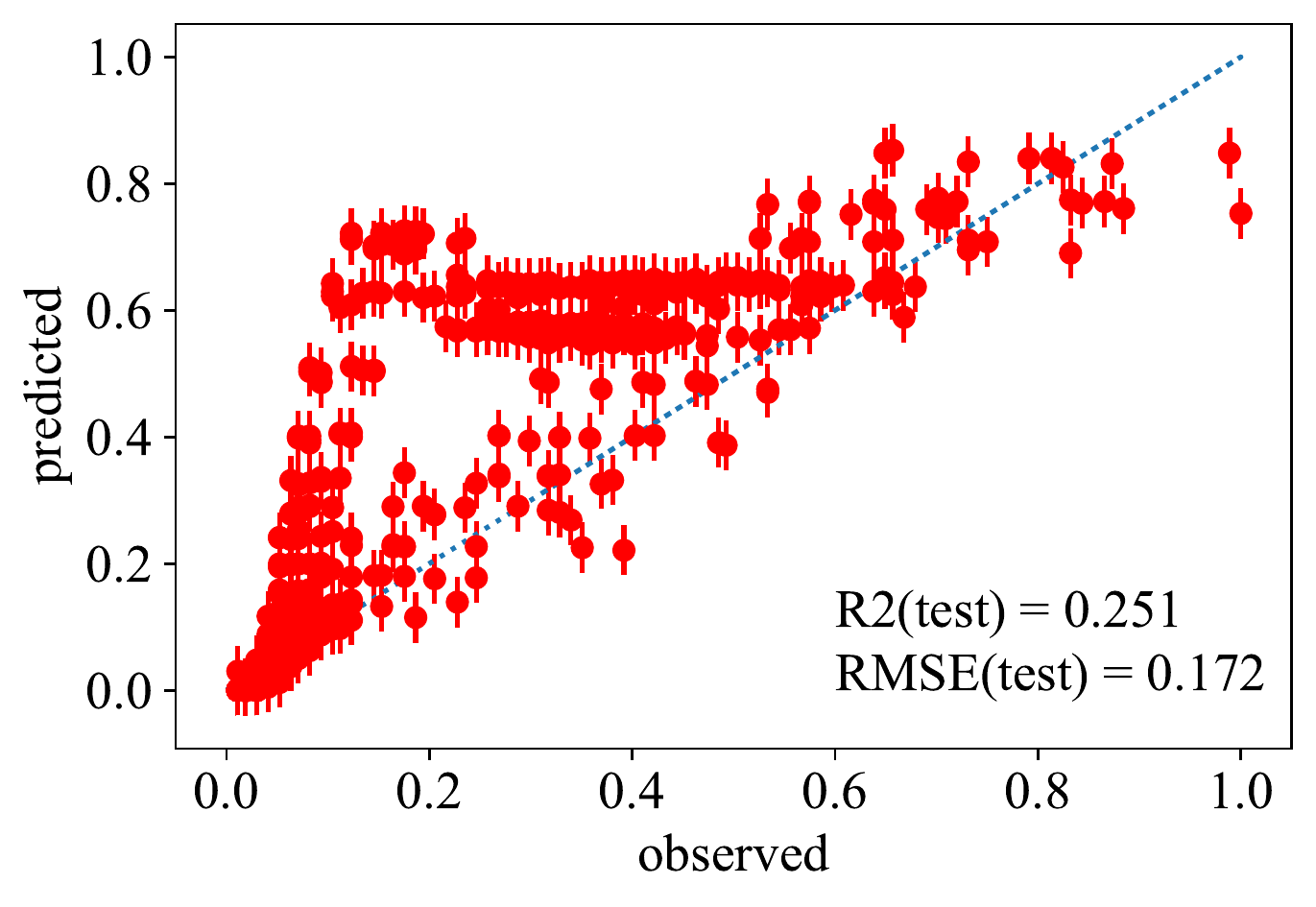} 
\caption{Validation of the  discovered model $g(T,H,\theta)$ on the testing data set, $t> 0.08$.}
\label{test_bhm_008}
\end{figure}

\begin{figure}[t]
\centering
\includegraphics[width=0.75\columnwidth]{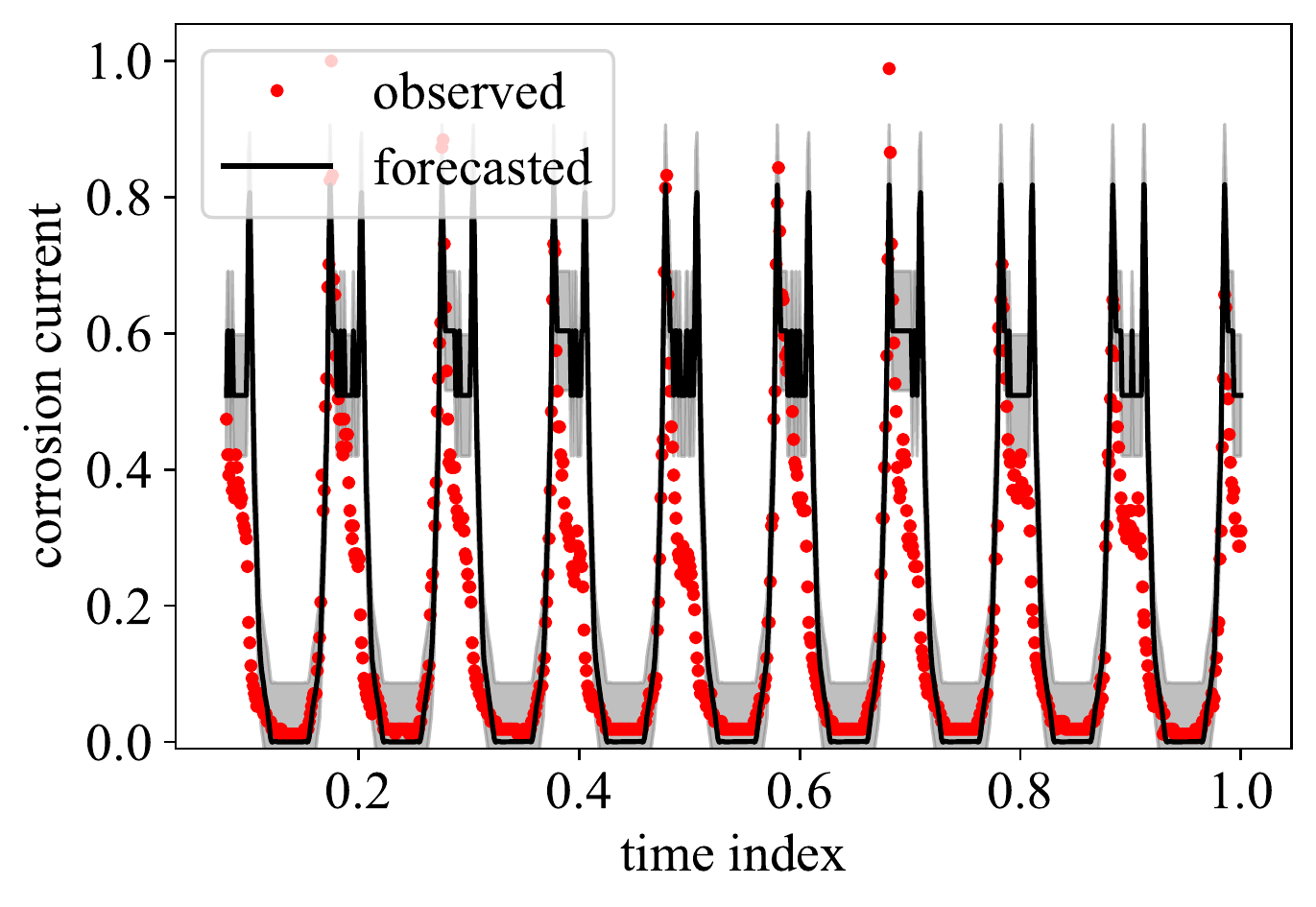} 
\caption{	The forecasting of the discovered model $g(T,H,\theta)$ for $t>0.08$. The confidence bounds are defined as ($\mu \pm \sigma$).}
\label{forecast_bhm_008}
\end{figure}

Next, the new discovered model $g(T,H,\theta)$ is calibrated using  using  GEBHM. The forecasting and validation results are shown in Fig.~(\ref{forecast_bhm_008}) and Fig.~(\ref{test_bhm_008}), respectively. The prediction of the current discovered model follows the trend of the observed data; however, shows an overestimation to the corrosion current between the peaks. Nevertheless, an improvement of  $239\%$ as shown in Fig.~(\ref{test_bhm_008}) can be achieved compared to the baseline model shown in Fig.~(\ref{test_deter_008}). The large improvement obtained here indicates that although the discovered model  has improved the performance by  $78\%$ (as shown previously), it is still missing some physics which can be captured by GEBHM based calibration.

\begin{table}[t]
\caption{Discovered correction terms to the  Butler-Volmer equation for the training data set $t\le 0.08$. Performance metrics are based on the testing data set $t>0.08$..}    
\label{DPD2}
\begin{center}
    \begin{tabular}{| l | l | l | l |  }
    \hline
BIC	& $R^2$			& $\delta(H)$ & V[\%]\\ \hline
-5620.2	&0.132	& $0.391\exp(H)$&	78 \\ \hline
-5609.6	&0.098	&	 $0.99H + 0.083$&	32 \\ \hline
-5608.8 &0.095 & $0.274H^2 + 0.509H + 0.293$ &28\\ \hline
    \end{tabular}
\end{center}
\end{table}


\subsubsection{DPD without a prior model:}
\begin{figure}[t]
\centering
\includegraphics[width=0.75\columnwidth]{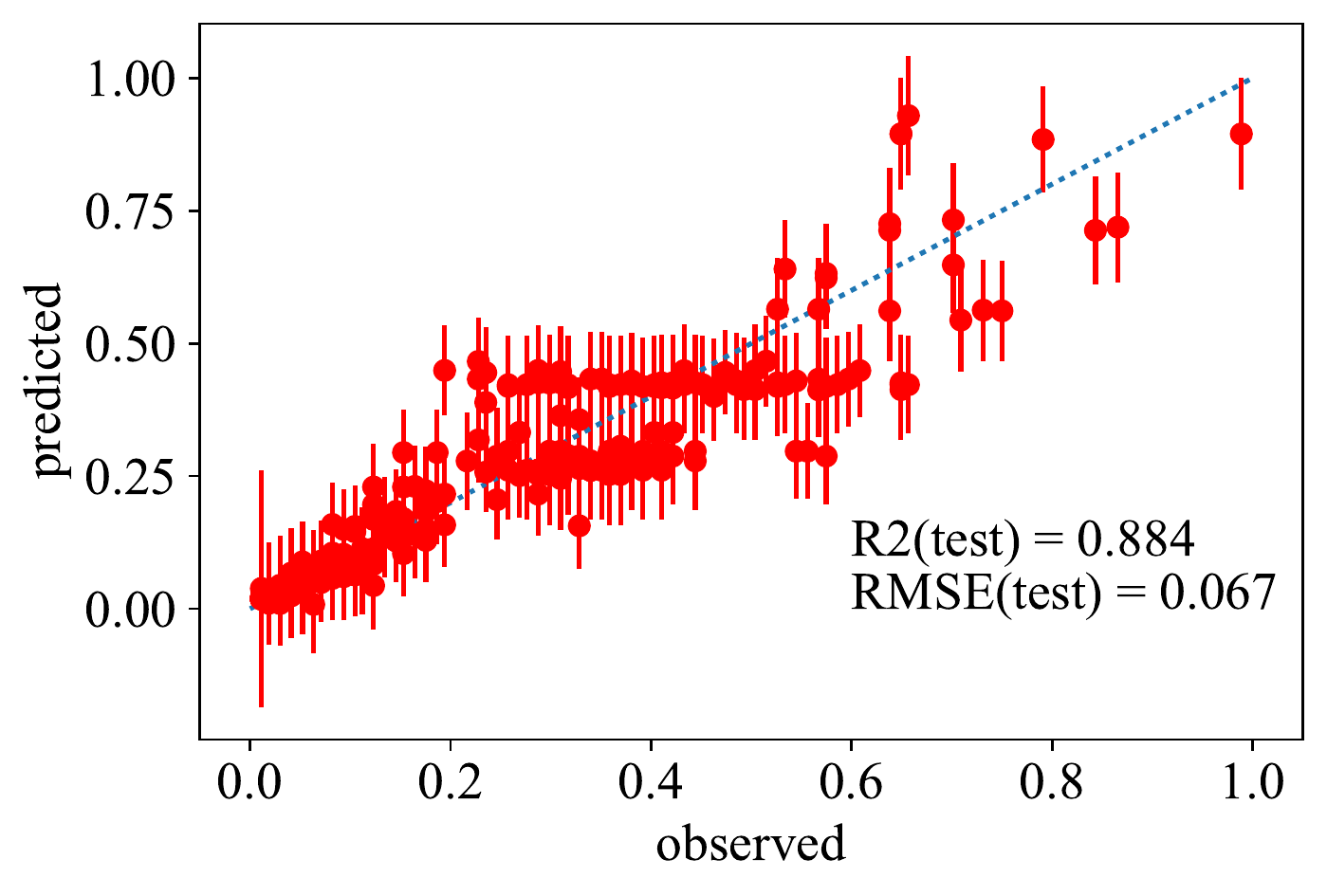} 
\caption{Validation of the  discovered standalone model $g(T,H,\theta)$ on the testing data set, $t> 0.5$.}
\label{test_bhm_noprior}
\end{figure}

\begin{figure}[t]
\centering
\includegraphics[width=0.75\columnwidth]{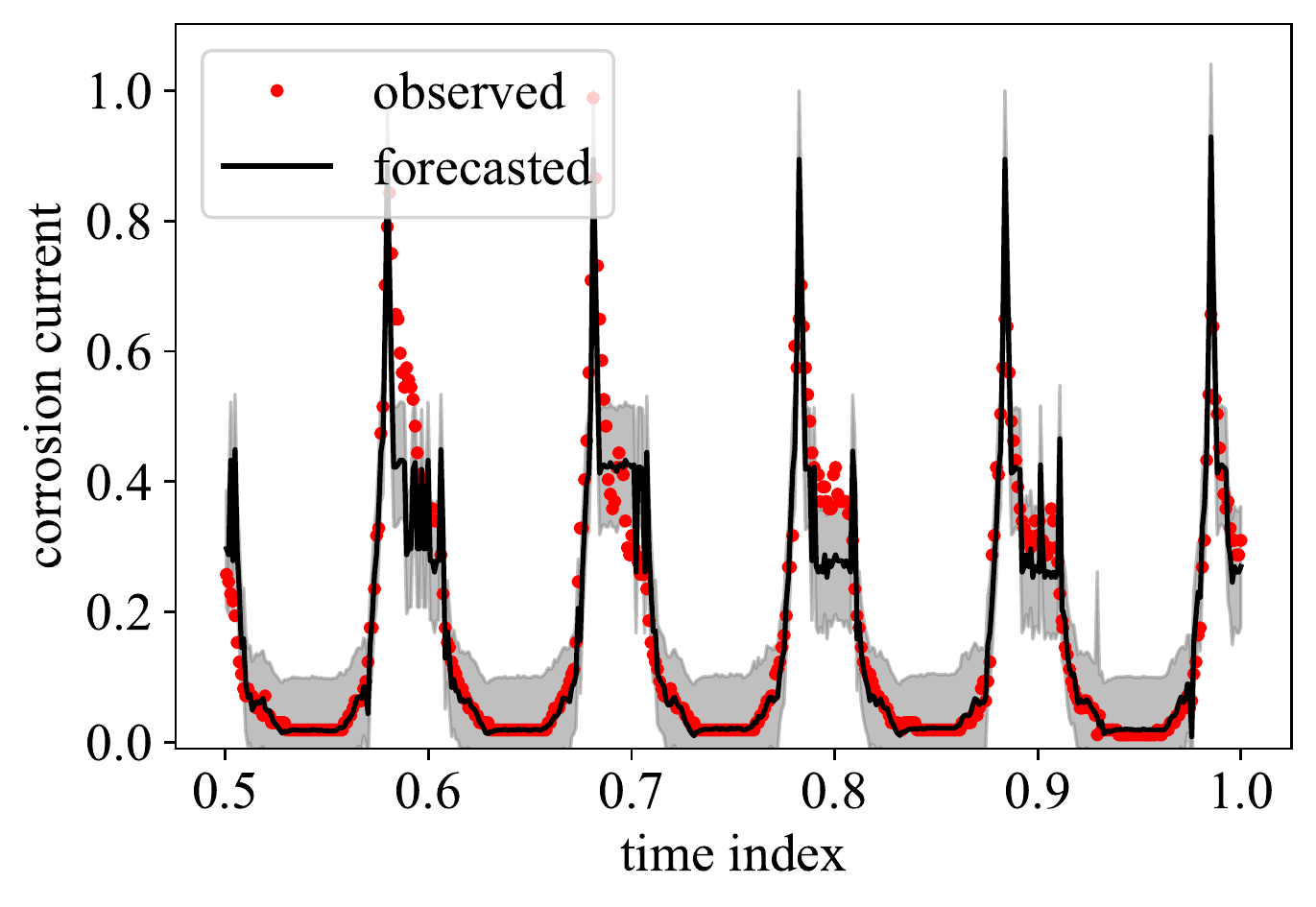} 
\caption{	The forecasting of the discovered standalone model $g(T,H,\theta)$ for $t>0.5$. The confidence bounds are defined as ($\mu \pm \sigma$).}
\label{forecast_bhm_noprior}
\end{figure}

Utilizing the  training data set $t\le 0.5$ only, Table~(\ref{DPD3}) shows some of the discovered standalone models without any prior physical model. Here the discovered model takes the form $g(T,H,\theta)=\delta(T,H,\theta_d)$. Large improvement can be achieved with more complicated model structure. These models explain the data in mathematical expressions, and they are ready for interpretation by the domain experts. Bayesian calibration of the discovered model $g(T,H,\theta)= H^2(\theta_0H^2 - \theta_1T)$ gives the validation results shown in Fig.~(\ref{test_bhm_noprior}), with an improvement of $28\%$. The forecasting performance of the discovered standalone model is shown in Fig.~(\ref{forecast_bhm_noprior}).

\begin{table}[t]
\caption{Discovered standalone models for the training data set $t\le 0.5$. Performance metrics are based on the testing data set $t>0.5$.}    
\label{DPD3}
\begin{center}
    \begin{tabular}{| l | l | l | l |  }
    \hline
BIC	& $R^2$			& $\delta(T,H)$ & V[\%]\\ \hline
-2235.6 &	0.862& 	$\begin{aligned}[c]  &0.537HT(H^2T + H - T + \\
&0.457(H - T)\exp(H(H + T)))\end{aligned}$	& 25 \\ \hline
-2220.5&	0.833&	$H^2(H^2 - 0.456T)$	& 21 \\ \hline
-2186.2&	0.825&	$\begin{aligned}[c]  
H^3&(-0.479T+\\ 
&0.479\exp(H) - 0.304) 
\end{aligned}$ &	19\\ \hline
-2195.5&	0.802&	$0.527H^4\exp(H - T)$	&16\\ \hline
    \end{tabular}
\end{center}
\end{table}

\section{Conclusions}
A Data-based Physics Discovery (DPD) framework for an automatic discovery of governing equations from observed data is presented. The framework can discover a correction term to an existing physical model in order to improve the prediction performance. The correction term can be a multiplicative or additive, and a function of the current input variable or new variable. In addition, the DPD can discover a  standalone model based on only the observed data, in case a prior model is not available. A Bayesian approach is utilized for the model calibration and validation under uncertainty. A real-world application in the aerospace industry is considered to show the practicality of the proposed framework.

\section{ Acknowledgments}
The authors thank the United States Air Force Research Laboratory for providing the corrosion data for public release under Distribution A as defined by the United States Department of Defense Instruction 5230.24. This material is based upon work supported by the Defense Advanced Research Projects Agency (DARPA) under Agreement No. HR00111990032. Approved for public release; distribution is unlimited.

\bibliography{ref}

\end{document}